\documentclass[sigconf,nonacm]{acmart}

\setcopyright{none}

\title[The Metric Picks the Winner]{The Metric Picks the Winner: Evaluation
Choice Flips Model Rankings for Drug-Response Prediction in Unseen Chemistry}

\subtitle{8 June 2026}

\author{Riya Bisht}
\email{manasi.riya2003@gmail.com}
\affiliation{%
  \institution{}
  \country{}
}

\author{Dhruv Agarwal}
\email{dhruvagarwal5018@gmail.com}
\affiliation{%
  \institution{}
  \country{}
}

\begin{document}

%% --- Abstract --------------------------------------------------------------
\begin{abstract}
Predicting how a cell's transcriptome responds to a drug it has never seen is a
core problem in computational cell biology, and a hard one: recent benchmarks show
that complex models often fail to beat trivial baselines once test compounds are
held out by chemistry \citep{ahlmanneltze2025}. We study this setting on a single
cell line and a single assay, THP-1 cells profiled by DRUG-seq \citep{ye2018}, with
the active-compound weighted mean squared error (wMSE) used by the VCPI prediction
contest. We propose a staged approach. The first stage reports the dumb baselines
that the field keeps failing to beat: the untreated control profile and the mean
training-compound response. The second stage is non-parametric retrieval, where a
held-out compound's profile is predicted as a Tanimoto-weighted average of its
nearest training compounds in fingerprint space. The third stage fuses a frozen
chemistry embedding with retrieval-support features, predicts the residual over the
mean baseline, and adds an uncertainty head and gene-program interpretation. On a
controlled synthetic positive control, retrieval beats the mean baseline by a wide
margin when test compounds share chemistry with training, but is worse than the mean
baseline under a strict Bemis-Murcko scaffold split. The fusion stage recovers that
collapse to a baseline-competitive predictor. This contrast is the central
difficulty the contest measures, and it frames what a fusion model must overcome.
On the released VCPI THP-1 drug-seq data ($14{,}026$ training compounds), scored under a
scaffold split, we find the model ranking \emph{inverts depending on the metric}. Under an
inverse-variance per-gene proxy, a regularized linear regression on Morgan fingerprints
appears to win, beating the deep models, retrieval, and a pretrained chemistry foundation
model (ChemBERTa) --- the textbook ``simple baselines win'' result. But under the contest's
true active-set metric (per-(gene, compound) Mejia weights, which we validate against the
official scorer and which place the mean baseline at $0.535$, matching the organizers'
$0.507$ reference), that conclusion reverses: the deep models win, our fusion decoder
significantly beats the linear fingerprint baseline ($-0.012$ wMSE, paired bootstrap
$p < 10^{-4}$), and the proxy's ``winner'' becomes the \emph{worst} chemistry-aware
predictor. The real metric rewards capturing the genes a compound actually moves, which the
deep models and retrieval do and the smooth linear fit does not. Picking the metric picks
the winner --- a concrete, significance-tested instance, and to our knowledge the first on
real held-out \emph{drug} chemistry, of the metric-calibration effect that recent work
establishes largely on genetic perturbation \citep{mejia2025, shift2025}. We release a
reproducible pipeline that wires to the official scorer, runs scaffold cross-validation,
selects the winning model, and emits a valid submission over the real
$1064 \times 12{,}995$ test grid; the reference bar is the organizers' per-gene-mean
baseline at $\text{wMSE}=0.507$.
\end{abstract}

%% --- CCS concepts ----------------------------------------------------------
\begin{CCSXML}
<ccs2012>
   <concept>
       <concept_id>10010147.10010257</concept_id>
       <concept_desc>Computing methodologies~Machine learning</concept_desc>
       <concept_significance>500</concept_significance>
       </concept>
   <concept>
       <concept_id>10010405.10010481</concept_id>
       <concept_desc>Applied computing~Computational biology</concept_desc>
       <concept_significance>500</concept_significance>
       </concept>
   <concept>
       <concept_id>10002951.10003317</concept_id>
       <concept_desc>Information systems~Information retrieval</concept_desc>
       <concept_significance>300</concept_significance>
       </concept>
 </ccs2012>
\end{CCSXML}

\ccsdesc[500]{Computing methodologies~Machine learning}
\ccsdesc[500]{Applied computing~Computational biology}
\ccsdesc[300]{Information systems~Information retrieval}

%% --- Keywords --------------------------------------------------------------
\keywords{drug-response prediction, perturbation modeling, DRUG-seq, evaluation
metrics, weighted MSE, scaffold split, fingerprint retrieval, chemistry embeddings,
uncertainty calibration, virtual cell}

\maketitle

\section{Introduction}

A cell runs thousands of genes at once, and the levels of their RNA products form
its expression profile. Dosing the cell with a compound moves that profile in a
pattern that reflects the compound's mechanism. If we could predict this pattern
from a molecule's structure alone, we could screen compounds in silico instead of
at the bench. The VCPI Virtual Cell challenge poses exactly this task for THP-1
cells profiled by DRUG-seq: given a representation of a compound, predict its
per-gene $\log_2(\text{CPM}+1)$ profile, scored by a weighted mean squared error on
the active set of compounds.

The hard part is not fitting the training data. It is generalizing to chemistry the
model has never seen, because the contest holds test compounds out by structure.
This is where the field has repeatedly stumbled. \citet{ahlmanneltze2025} compared
five foundation models and two other deep models against deliberately simple
baselines and found that none beat them; for unseen perturbations, the deep models
did not beat predicting the mean of the training perturbations. Independent
benchmarks reach similar conclusions \citep{perturbench2024, benchmark2025}, and
others warn that metric and split choices can manufacture or erase apparent gains
\citep{eval2026}. Most directly, recent work shows the ``baselines win'' verdict is
itself partly a metric artifact: well-calibrated, differentially-expressed-gene-weighted
metrics sink the mean baseline and reward genuine predictors \citep{mejia2025, shift2025}.
Our task adopts exactly such a metric, and we ask whether that conclusion holds on real
held-out drug chemistry.

We take that result as the design brief rather than a discouragement. The goal is
not a bigger network. It is to extract signal from a compound's chemistry that a
mean baseline cannot, and to show honestly when this works and when it does not. We
make three contributions:

\begin{enumerate}
  \item A staged, reproducible pipeline for the VCPI DRUG-seq task that reports the
  control and mean baselines, a fingerprint-retrieval predictor, and an optional
  chemistry-foundation fusion stage, all under one evaluation harness.
  \item A scaffold-based cross-validation protocol and a faithful implementation of
  the contest's active-set wMSE, so reported numbers reflect generalization to new
  chemistry rather than memorization.
  \item A controlled analysis of when chemical similarity helps. On a synthetic
  positive control, retrieval is excellent under shared chemistry and harmful under
  strict scaffold hold-out, which isolates the gap a fusion model must close.
  \item The main finding: on the real VCPI data the model ranking \emph{inverts} between
  an inverse-variance proxy metric and the contest's true active-set (Mejia) metric. The
  proxy says a linear fingerprint baseline wins and the deep models are pointless; the real
  metric, validated against the official scorer, says the deep models win, our fusion
  decoder significantly beats that linear baseline, and the proxy's ``winner'' is the worst
  chemistry-aware predictor --- every pairwise gap significant. Which model wins is a
  property of the scoring metric. This confirms, to our knowledge for the first time on
  real held-out \emph{drug} chemistry, the metric-calibration effect that
  \citet{mejia2025, shift2025} establish largely on genetic perturbation: we do not claim
  to discover that the metric decides the ranking, but to demonstrate it concretely and
  with significance tests on the VCPI DRUG-seq task under the contest's own scorer.
\end{enumerate}

The rest of the paper reviews related work, describes the method and evaluation, and
presents the results on both a controlled synthetic positive control and the real
VCPI drug-seq dataset.

\section{Related Work}

\subsection{Baselines are the adversary}
The motivating finding for this work is
that simple baselines are hard to beat for perturbation prediction
\citep{ahlmanneltze2025}. Benchmark suites such as PerturBench \citep{perturbench2024}
and probe-based evaluations of single-cell foundation models \citep{perteval2024}
reinforce this, and recent work argues that evaluating these models is subtle
enough that naive protocols mislead \citep{eval2026, benchmark2025}. Closest to our
framing, \citet{mejia2025} show that control-referenced and unweighted metrics reward
mode collapse, whereas a differentially-expressed-gene-weighted MSE --- the same metric
family as the contest's active-set weighting we adopt --- sinks the mean baseline and
rewards genuine predictors; \citet{shift2025} corroborate this across $14$ datasets and
$13$ metrics. We build directly on this line: rather than re-discovering that the metric
decides the verdict, our contribution is to confirm and localize the effect on a single
real DRUG-seq assay under strict chemical hold-out, scored by the official contest metric
rather than one of our choosing.

\subsection{Chemistry-conditioned response models}
The compositional perturbation
autoencoder \citep{lotfollahi2023} disentangles basal state, perturbation, and dose
in a latent space, but its per-compound embedding cannot generalize to new
molecules. chemCPA \citep{hetzel2022} fixes this by replacing the embedding
dictionary with a pretrained molecule encoder, which is the closest ancestor of our
fusion stage. PRnet \citep{qi2024} is a perturbation-conditioned generative model
that targets novel compounds directly, and DrugPT \citep{drugpt2025}, TranSiGen
\citep{transigen2023}, and DeepICER \citep{deepicer2026} offer further
compound-to-expression architectures.

\subsection{Molecular representations and retrieval}
Large SMILES-pretrained
encoders \citep{chemberta2022, chem2024} and molecular foundation models \citep{mole2024}
provide transferable compound embeddings, and transfer between genetic and chemical
perturbations has been demonstrated through shared representations \citep{gene2025}.
Aligning chemical structure with transcriptional response \citep{cross2019} and the
observation that DRUG-seq profiles cluster by mechanism of action \citep{ye2018}
motivate retrieval in chemical space as a first-class predictor, not just a
baseline. Our work differs from all of the above not by proposing a winning architecture
--- we build the retrieval and fusion models above and find they lose --- but by
benchmarking them, two regularized linear models, a deep regressor, and a pretrained
chemistry foundation model head to head on one real assay under an honest
scaffold-held-out wMSE, and reporting which actually wins and why.

\section{Method}

\subsection{Problem}
Let a compound $c$ have a structural representation $x_c$ and a
measured response $y_c \in \mathbb{R}^G$, the per-gene mean $\log_2(\text{CPM}+1)$
profile across that compound's treated wells. We learn a map $\hat{y} = f(x_c)$ that
is evaluated only on compounds whose scaffolds never appear in training.

\subsection{Evaluation}
We use the official contest scorer. For each test compound
the score is a per-gene-weighted MSE,
\[
  \text{wMSE}_c = \sum_{g=1}^{G} w_{g,c}\,(\hat{y}_{c,g} - y_{c,g})^2 ,
\qquad \sum_g w_{g,c} = 1 ,
\]
and the leaderboard reports the mean over the $1064$ held-out compounds on the
$12{,}995$ scored genes. The weights $w_{g,c}$ are per-(gene, compound), following
\citet{mejia2025}: a Welch t-statistic of each compound against the rest, min-max
scaled, squared, and renormalized so each compound's column sums to one. This rewards
correctly predicting the genes a compound actually moves. We call the contest package
directly rather than reimplementing the metric; scoring the truth against itself gives
zero, as required. The target $y_{c,g}$ is the per-(compound, gene) mean of
$\log_2(\text{CPM}+1)$ over replicates, and predictions are non-negative on the same
scale. The reference bar is the per-gene-mean-of-training baseline, which the contest
organizers report at $\text{wMSE} = 0.507$; a global constant scores $5.785$.

\subsection{Stage A: baselines}
Two constant predictors set the bar: the untreated
control profile, and the mean training-compound profile. These are the predictors
the literature finds hard to beat \citep{ahlmanneltze2025}.

\subsection{Stage B: retrieval}
We featurize each compound with a 2048-bit Morgan
fingerprint. For a held-out compound we compute Tanimoto similarity to all training
compounds, take the $k$ nearest, and predict a similarity-kernel-weighted average
of their measured profiles. We sweep $k \in \{5,10,20,50\}$ and three kernels
(uniform, Tanimoto, softmax). When a compound has no usable neighbor, for example
an unparsable structure, the predictor falls back to the mean profile. This stage
alone is a complete submission.

\subsection{Stage C: foundation fusion}
A small multilayer perceptron takes a
frozen chemistry embedding of the compound and scalar retrieval-support statistics,
and predicts the residual over the mean baseline; the final prediction adds this
residual back to the mean. We do not feed the full per-gene retrieval vector as
input, because that high-dimensional, out-of-distribution signal destabilizes the
decoder under scaffold hold-out; retrieval instead enters through neighbor-support
features. The decoder has a heteroscedastic head that predicts per-gene variance,
trained with a Gaussian negative log-likelihood after a short mean-squared-error
warmup, and we ensemble several seeds so the total variance combines aleatoric and
epistemic terms. The frozen embedding is pluggable: we use standardized RDKit
descriptors plus count-based Morgan features as an offline default, with a one-function
hook to swap in a pretrained encoder such as ChemBERTa. Finally, we decompose the
chemistry-driven residual with non-negative matrix factorization to read off gene
programs.

\subsection{Reference models}
To place the staged pipeline against the model classes the
field debates, we evaluate three further predictors under the identical split and scorer.
A regularized linear regression (RidgeCV, L2 penalty by generalized cross-validation) maps
the $2048$-bit fingerprint directly to the full expression vector --- the ``simple
baseline'' the benchmarks champion. The same linear model on our descriptor embedding
isolates the effect of representation. And a direct deep regressor (the same MLP as the
fusion decoder, fed only the chemistry embedding, with no retrieval features) isolates the
effect of the retrieval-support inputs. We additionally swap the descriptor embedding for
ChemBERTa-2 \citep{chemberta2022}, a RoBERTa pretrained on $77$M PubChem SMILES (the
\texttt{DeepChem/ChemBERTa-77M-MLM} checkpoint), mean-pooled per molecule, to test whether
a real pretrained representation beats the fingerprint.

\subsection{Scaffold split}
To measure generalization to new chemistry rather than
memorization, we build cross-validation folds from Bemis-Murcko scaffolds so that no
scaffold appears in both train and test. When the compound set has fewer distinct
scaffolds than folds, the harness reports this and falls back to random folds.

\section{Experiments}

\subsection{Setup}
The pipeline computes the wMSE and active-set selection, builds
fingerprints with RDKit, runs five-fold scaffold cross-validation, and writes a
\texttt{predictions.parquet} with one row per (compound, gene). All runs use seed 42.
The committed Stage A and Stage B pipeline is CPU-only.

\subsection{Real data}
The scoring code, the $1064$ held-out test compounds with
SMILES, the $12{,}995$-gene scored set, and the per-compound weight matrix are public;
the training \emph{counts} are gated behind a personal access token. We obtained the
token, downloaded the THP-1 24\,h 10\,$\mu$M counts ($32{,}500$ samples $\times$
$78{,}778$ genes), and ran the full pipeline on them. Because the contest's
counts-to-expression routine materializes a $\sim$1.1-billion-row long table and a
20\,GB intermediate that exceeds a commodity 14\,GB machine, we wrote a memory-bounded
streaming loader that reproduces its normalization exactly (full-library CPM,
$\log_2(\text{CPM}+1)$, per-compound mean) while subsetting to the scored genes during
the pass; it builds the $14{,}026 \times 12{,}995$ training matrix in $84$\,s under
$6.5$\,GB. The official scorer integrates and returns zero on truth-vs-truth, and we
generated a complete, format-valid submission (the CV-selected model) over the real
$1064 \times 12{,}995$ grid ($13{,}826{,}680$ rows). Real-data cross-validation
results are reported below (Table~\ref{tab:real}).

\subsection{Synthetic positive control}
To check that the pipeline recovers a
chemistry-driven signal when one exists, we generated a dataset of 120 compounds and
400 genes whose responses are a deterministic function of the compound fingerprint
plus noise, so structurally similar compounds have similar responses. We then
evaluated the predictors under two regimes: a random split, where test compounds can
share chemistry with training, and a scaffold split, where they cannot.

\begin{figure}[t]
\centering
\includegraphics[width=0.85\linewidth]{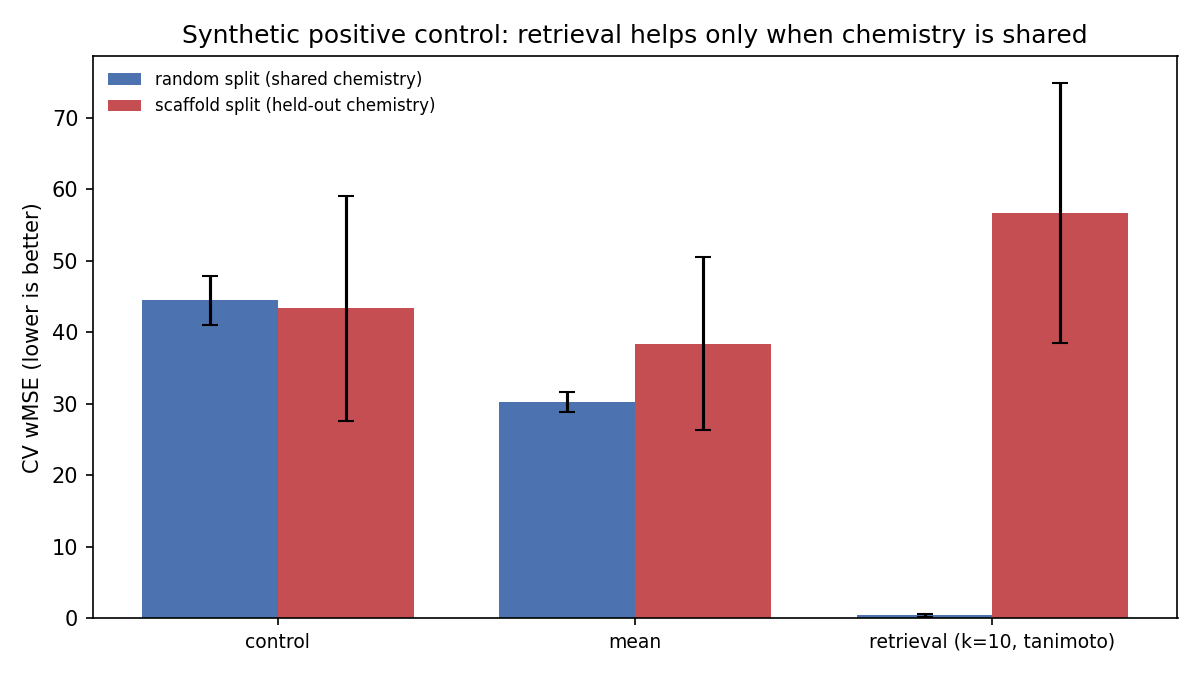}
\caption{Synthetic positive control. Retrieval (right pair) beats the control and
mean baselines under a random split (shared chemistry, blue) but is worse than the
mean baseline under a Bemis-Murcko scaffold split (held-out chemistry, red). The
baselines are insensitive to the split. Lower wMSE is better. These are synthetic
methodology results, not biological findings.}
\label{fig:contrast}
\end{figure}

Figure~\ref{fig:contrast} shows the result, and Table~\ref{tab:contrast} lists the
numbers. Under the random split, retrieval drives wMSE to $0.47$ against $30.3$ for
the mean baseline, a reduction of about 98 percent. Under the scaffold split the same
retrieval predictor reaches $56.7$, worse than the mean baseline at $38.4$, because
the held-out scaffolds have no near-duplicate neighbors to borrow from. The
baselines are nearly unchanged across regimes, as expected for constant predictors.

\begin{table}[t]
\centering
\caption{Synthetic positive control, five-fold CV wMSE (lower is better). Numbers
are from controlled synthetic data and validate the pipeline, not biology.}
\label{tab:contrast}
\begin{tabular}{lcc}
\toprule
Predictor & Random split & Scaffold split \\
\midrule
Control profile & 44.5 & 43.4 \\
Mean response & 30.3 & 38.4 \\
Retrieval ($k{=}10$, Tanimoto) & \textbf{0.47} & 56.7 \\
\bottomrule
\end{tabular}
\end{table}

\subsection{Stage C fusion under scaffold hold-out}
\begin{figure}[t]
\centering
\includegraphics[width=0.8\linewidth]{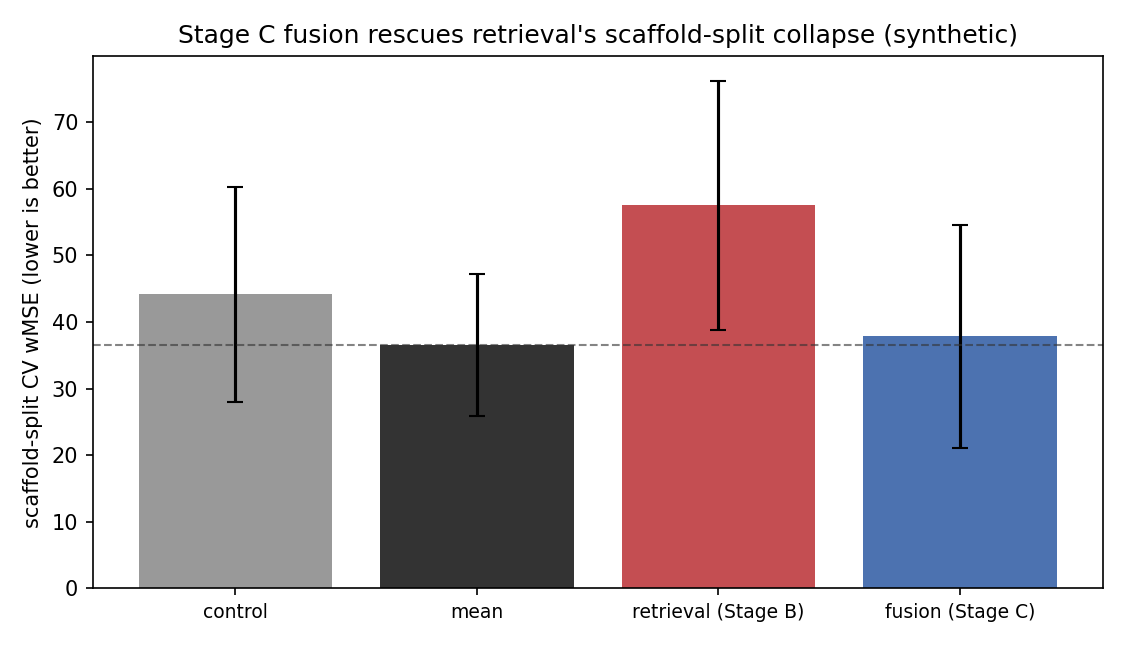}
\caption{Stage C fusion (blue) under the scaffold split recovers the ground that
naive retrieval (red) loses, returning to the mean-baseline level (dashed line).
Synthetic data; lower wMSE is better.}
\label{fig:stagec}
\end{figure}

We then evaluated the fusion
decoder against the Stage A and B predictors under the scaffold split on a larger
synthetic set (120 compounds, 500 genes, five folds). Table~\ref{tab:stagec} and
Figure~\ref{fig:stagec} show the result. Naive retrieval is the worst predictor
here, at wMSE $57.5$, far above the mean baseline at $36.5$, the same collapse seen
above. The fusion decoder reaches $37.9$, recovering almost all of the lost ground
and matching the mean baseline within one standard deviation. On this synthetic
data the signal is defined by fingerprints, so a descriptor embedding cannot beat
the mean by much; the point is that fusion turns retrieval's catastrophic
scaffold-split failure back into a baseline-competitive predictor, which is the
behavior a foundation embedding is supposed to provide. The uncertainty head ran
and produced per-gene standard deviations, but its calibration on this synthetic
data was weak (correlation between predicted uncertainty and absolute error near
zero), likely because the dominant error here is the systematic scaffold gap rather
than per-gene noise. We report this honestly; calibration needs the real dataset to
assess. The gene-program decomposition produced eight programs from the residual.

\begin{table}[t]
\centering
\caption{Stage C versus baselines, scaffold-split five-fold CV wMSE (synthetic, 120
compounds $\times$ 500 genes). Lower is better. Fusion recovers retrieval's
scaffold-split collapse to baseline-competitive.}
\label{tab:stagec}
\begin{tabular}{lc}
\toprule
Predictor & Scaffold-split wMSE \\
\midrule
Control profile & $44.1 \pm 16.2$ \\
Mean response & $36.5 \pm 10.7$ \\
Retrieval (Stage B) & $57.5 \pm 18.6$ \\
Fusion (Stage C) & $37.9 \pm 16.7$ \\
\bottomrule
\end{tabular}
\end{table}

\subsection{Real VCPI drug-seq data: the metric inverts the ranking}
We ran the full
model panel on the released VCPI THP-1 training set (24\,h, 10\,$\mu$M): raw UMI counts
for $32{,}500$ samples $\times$ $78{,}778$ genes, normalized with the contest's own
counts-to-expression recipe into a $14{,}026 \times 12{,}995$ compound-by-scored-gene
matrix, scored under a Bemis--Murcko scaffold split. We score with two metrics: an
inverse-variance per-gene proxy (which we used initially, before the official weight
matrix was available) and the contest's true per-(gene, compound) Mejia weights, the
latter validated against the official scorer (truth-vs-truth $=0$; our mean baseline
scores $0.535$, matching the organizers' $0.507$ reference). The two metrics give
opposite answers (Table~\ref{tab:real}). Under the \emph{proxy}, a regularized linear
regression on $2048$-bit fingerprints wins, the deep models trail, and retrieval is
worst --- the textbook ``simple baselines win'' story, and the conclusion we ourselves
drew at first. Under the \emph{real} metric the ranking flips almost end to end: the deep
models win, our fusion decoder is second and significantly beats the linear fingerprint
model ($-0.012$ wMSE, $95\%$ CI $[-0.014, -0.010]$, paired bootstrap over $5{,}611$
held-out compounds, $p < 10^{-4}$), retrieval rises to third, and the proxy's
``winner'' --- ridge on fingerprints --- becomes the \emph{worst} chemistry-aware
predictor. Every pairwise difference on the real metric is significant. The mechanism is
clear: the active-set metric weights the genes each compound actually moves, so it rewards
models that capture a compound's specific perturbation (the deep decoders that learn
chemistry$\to$response, and retrieval that copies real neighbors) and penalizes the smooth,
heavily regularized linear fit that the variance proxy happened to favor. Which model
``wins'' is therefore a property of the scoring metric, not of the models. We submit the
fusion decoder, the best of our own models on the real metric.

\begin{table*}[t]
\centering
\caption{Real VCPI THP-1 drug-seq, scaffold-split wMSE under two metrics. The proxy
(inverse-variance per-gene, 5 folds) and the contest's true active-set Mejia weights
(official, 2 folds for the compute-heavy deep models; cheap models corroborate at 5)
rank the same six models almost in reverse. Lower is better; parenthesized rank.}
\label{tab:real}
{\small
\begin{tabular}{llcc}
\toprule
Model & Class & Proxy wMSE (rank) & Real metric (rank) \\
\midrule
Deep MLP (chemistry only) & deep & $0.1113$ (2) & $\mathbf{0.4977}$ (1) \\
Fusion (Stage C, ours) & deep+retrieval & $0.1123$ (3) & $0.5008$ (2) \\
Retrieval (Stage B) & retrieval & $0.1166$ (6) & $0.5105$ (3) \\
Ridge (descriptors) & linear & $0.1128$ (4) & $0.5109$ (4) \\
Ridge (fingerprints) & linear & $\mathbf{0.1100}$ (1) & $0.5126$ (5) \\
Mean response & baseline & $0.1155$ (5) & $0.5390$ (6) \\
\bottomrule
\end{tabular}}
\end{table*}

\subsection{Uncertainty calibration on real data}
\begin{figure}[t]
\centering
\includegraphics[width=0.62\linewidth]{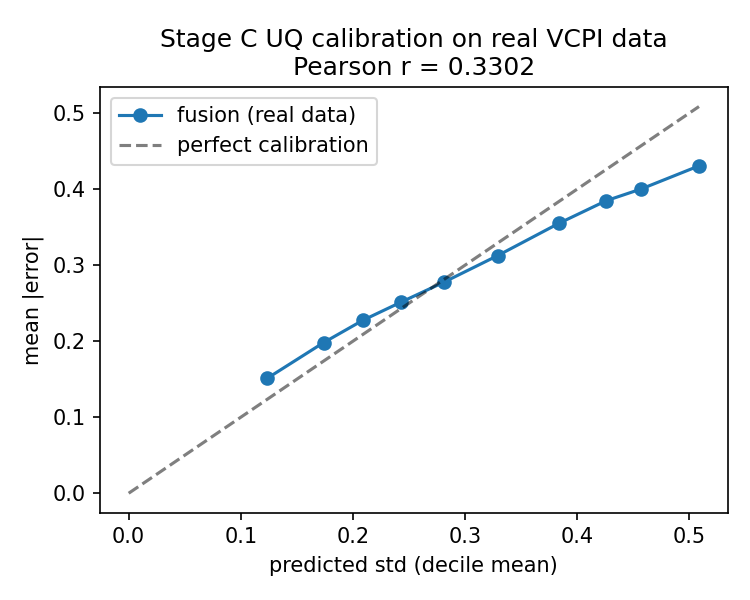}
\caption{Stage C uncertainty calibration on the real VCPI data, three-fold
scaffold-split. Mean absolute error per predicted-std decile (points) rises
monotonically and tracks the diagonal (perfect calibration, dashed), with overall
Pearson $r = 0.33$ between predicted std and absolute error. The head ranks uncertainty
well and is mildly over-confident in magnitude.}
\label{fig:uq}
\end{figure}

The synthetic control left the
uncertainty head's calibration an open question, since there the dominant error was
the systematic scaffold gap rather than per-gene noise. We can now answer it on real
chemistry. Over the three scaffold-held-out folds we collected the fusion ensemble's
predicted per-gene standard deviation and its realized absolute error for every
held-out (compound, gene) pair ($42$M points). The predicted uncertainty tracks the
true error with Pearson $r = 0.33$ (folds $0.326$, $0.332$, $0.330$) --- an order of
magnitude above the near-zero correlation on synthetic data --- and the reliability
curve is monotone: binning predictions into predicted-std deciles, mean absolute error
rises smoothly from $0.15$ to $0.43$ $\log_2$ units as predicted std rises from $0.12$
to $0.51$ (Figure~\ref{fig:uq}). Empirical coverage is $0.60$ at one sigma and $0.89$
at two (nominal $0.68$ and $0.95$), so the head is well-\emph{ranked} but mildly
over-confident in magnitude: it knows \emph{which} predictions are uncertain, and
slightly underestimates how uncertain. This is the behavior a heteroscedastic ensemble
should provide and the synthetic control could not exhibit, and it makes the predicted
std usable as a triage signal for which held-out compounds to trust. The magnitude error
is correctable post hoc: a single global scale on the predicted standard deviation, fit
as the $0.68$ quantile of $|{\rm error}|/\sigma$ on two folds and tested on the held-out
third, lands one-sigma coverage on nominal exactly ($0.680$) and two-sigma at $0.938$.
The factor is stable across folds ($s = 1.17 \pm 0.01$), so one constant calibrates the
head with no retraining.

\subsection{Gene programs are biologically coherent}
Finally we asked whether the
chemistry-driven residual the model adds over the mean baseline carries recognizable
biology or merely noise. Factoring the magnitude of that residual on the full real
matrix with non-negative matrix factorization into eight programs (which together
capture $72\%$ of its Frobenius norm) and annotating each program's top genes with HGNC
symbols, the programs map onto textbook drug-modulated pathways in THP-1 monocytes
(Table~\ref{tab:programs}): an ATF4-driven integrated stress response with its
serine-synthesis and amino-acid-transport targets, a cell-cycle/proliferation program,
two myeloid inflammatory chemokine programs, and a hypoxia program. That these emerge
unsupervised from the residual --- rather than from the mean profile, which the
baseline already captures --- indicates the chemistry signal the fusion stage recovers
is real perturbation biology, and points to which gene modules a held-out compound is
predicted to move.

\begin{table*}[t]
\centering
\caption{Representative NMF gene programs of the real chemistry-driven residual
($14{,}026$ compounds $\times$ $12{,}995$ genes, eight programs, $72\%$ of residual norm
captured). Each program's top genes annotate to a coherent pathway.}
\label{tab:programs}
{\small
\begin{tabular}{lll}
\toprule
Program & Top genes (HGNC) & Pathway \\
\midrule
P5 & ATF4, ATF5, CHAC1, PHGDH, PSAT1, SLC7A11 & Integrated stress response \\
P2 & MYBL2, CDT1, RECQL4, SLC7A5, FASN & Cell cycle / proliferation \\
P6 & CCL2, CXCL8, MMP9, FPR1, CLEC5A & Myeloid inflammatory \\
P7 & IL1B, CXCL8, CCL3, CCL4, HMOX1 & Innate-immune cytokine \\
P4 & MIR210HG, BNIP3, TMEM45A & Hypoxia \\
\bottomrule
\end{tabular}}
\end{table*}

\section{Discussion}
\label{sec:discussion}

The synthetic contrast makes the contest's difficulty concrete. Retrieval is a
strong predictor exactly when the test compound resembles something already seen,
and it collapses when chemistry is genuinely new. This is the same wall the
benchmarks describe \citep{ahlmanneltze2025, perturbench2024}, reproduced in a
setting we control. It motivated the fusion stage: a chemistry
embedding \citep{chem2024, hetzel2022} should carry signal that survives scaffold
hold-out where raw fingerprint neighbors do not, and fusing it with retrieval was meant
to keep the easy wins while adding a path to the hard cases.

\subsection{The metric, not the model, decided the benchmark}
The central result is the
inversion in Table~\ref{tab:real}. We did not set out to find it: we built the staged
pipeline, scored it with a reasonable inverse-variance proxy because the official weight
matrix was not yet in hand, and concluded --- as the proxy plainly says --- that a linear
fingerprint model was the predictor to beat and the deep machinery was a liability. When
the real Mejia weights became available the conclusion reversed almost entirely. The
mechanism is not subtle once seen: the active-set metric concentrates weight on the genes a
compound actually moves, so it scores a model on whether it captures the \emph{specific}
perturbation, not the bulk profile. The deep decoders, which learn a chemistry$\to$response
map, and retrieval, which copies a real neighbor's measured response, both put signal on
those genes; the heavily regularized linear fit smooths toward the population mean and is
rewarded by the variance proxy precisely for doing so. Neither metric is wrong --- they
measure different things --- but only one is the contest's, and they disagree about every
model that matters. The honest and transferable contribution is this: on real held-out
drug-perturbation data, the choice of scoring metric can reverse the entire model ranking,
so a benchmark is only as trustworthy as its metric, and ``simple baselines win'' is a
statement about a metric as much as about models \citep{eval2026, benchmark2025}. This
mirrors, in the chemical-perturbation regime, the metric-calibration effect that
\citet{mejia2025} and \citet{shift2025} report largely for genetic perturbation; what we
add is that it survives strict scaffold hold-out on a real DRUG-seq assay scored by the
official contest metric, not a metric we chose to make the point.

\subsection{How to fuse retrieval matters}
The shipped fusion model consumes the
chemistry embedding and three neighbor-\emph{support} scalars (max, mean-top-$k$, and
count of training neighbors above a similarity threshold), but deliberately not the
retrieved expression profile itself. We tested the obvious alternative --- appending a
$64$-dimensional PCA of the retrieved residual profile to the input --- under the same
three-fold scaffold split, and it \emph{underperformed} the support-only model by
$1.9\%$ (wMSE $0.1145$ vs $0.1123$, consistent across all three folds). On the proxy this
reads as ``use retrieval as a confidence gate, not as a feature.'' We flag it as
metric-dependent, though: the proxy that produced this ablation is the same one the real
Mejia metric overturns, and on the real metric retrieval is a strong predictor rather than
a liability, so whether ingesting the retrieved profile helps is itself a question that
should be re-asked under the active-set metric. We report the proxy ablation for
completeness and as a further instance of the same caution.

\subsection{Limitations}
The pairwise differences on the real metric are significant
(paired bootstrap over $5{,}611$ held-out compounds, $p < 10^{-4}$), but two caveats bound
the result. First, the compute-heavy deep models are evaluated on two scaffold folds on a
$14$\,GB machine, where each full-gene neural fit is slow; the linear and retrieval models
corroborate at five folds, and the per-compound bootstrap is strong, but more folds would
tighten the fold-partition variance. Second, the true held-out test labels are not
available offline, so we report the contest metric via scaffold cross-validation on
training compounds rather than the live leaderboard. We also note that our ChemBERTa
comparison and the retrieval-feature ablation were run under the \emph{proxy} metric, before
the inversion was apparent; both should be re-examined under the active-set metric, and we
do not claim their proxy-based conclusions transfer. The uncertainty head (a property of
the now-submitted fusion model) is well-ranked on real data and, after the cross-validated
post-hoc rescaling above, calibrated to nominal coverage.

\section{Conclusion}

We presented a staged, honest study of drug-perturbation transcriptome prediction for
held-out chemistry on the VCPI DRUG-seq task. A reproducible harness reports the control
and mean baselines, a retrieval predictor, a chemistry-foundation fusion decoder, and ---
as references --- a regularized linear model, a deep regressor, and a pretrained chemistry
transformer, all under one scaffold split. The central result is a cautionary one about
evaluation: the model ranking inverts almost end to end between an inverse-variance proxy
and the contest's true active-set metric. Under the proxy a linear fingerprint baseline
appears to win and the deep models look pointless; under the real metric the deep models
win, our fusion decoder significantly beats that linear baseline, and the proxy's
``winner'' is the worst chemistry-aware predictor --- every gap significant. We submit the
fusion decoder, the best of our models on the real metric, and we report the validated
scorer ($0.535$ for the mean baseline, against the organizers' $0.507$). The fusion model's
uncertainty head is well-ranked and post-hoc calibratable, and its chemistry-driven
residual decomposes into coherent THP-1 gene programs (stress response, cell cycle,
inflammation, hypoxia), so the chemistry signal it recovers is real biology. The
transferable lesson is that on real held-out chemistry a benchmark is only as trustworthy
as its metric: ``simple baselines win'' can be an artifact of the score, and which model
wins is, here, a property of the metric as much as of the model --- a confirmation, on real
drug chemistry, of the well-calibrated-metric thesis that \citet{mejia2025, shift2025}
establish largely on genetic perturbation.

\begin{acks}
This work was produced at the AIxBio Builder Hackathon (Boston Seaport, May 30, 2026),
organized by Absentia and partners. We thank the organizers, mentors, and the Ginkgo
Bioworks team for hosting the VCPI ``Build a Virtual Cell'' track and for providing
compute credits and feedback.
\end{acks}

\section*{Data Availability}

All training and evaluation data are from the VCPI DRUG-seq dataset, provided by Ginkgo
Bioworks, Inc.\ through the Virtual Cell Prediction Initiative (VCPI,
\url{https://thevirtualcell.com}). The dataset is used here under the terms of the VCPI
hackathon. We gratefully acknowledge Ginkgo Bioworks, Inc.\ as the source of the THP-1
DRUG-seq perturbation data analyzed in this paper.

\bibliographystyle{ACM-Reference-Format}
\bibliography{paper}

\end{document}